\documentclass{article}
\usepackage{amsmath,graphicx}
\usepackage[utf8]{inputenc}
\usepackage{amssymb}
\usepackage{booktabs}
\usepackage{url}
\usepackage{idsiatechrep}

\title{Fast Image Scanning with\\ Deep Max-Pooling Convolutional Neural Networks}
\author{
Alessandro Giusti\\
Dan C.~Cire\c{s}an\\
Jonathan Masci\\
Luca M.~Gambardella\\
J\"{u}rgen Schmidhuber
}
\reportnumber{01-13}

\begin{document}
\makecover
%
\maketitle
\begin{abstract}
Deep Neural Networks now excel at image classification, detection and segmentation.  When used to scan images by means of a sliding window, however, their high computational complexity can bring even the most powerful hardware to its knees.  We show how dynamic programming can speedup the process by orders of magnitude, even when max-pooling layers are present.
\end{abstract}

\section{Introduction}


Deep Max-Pooling Convolutional Neural Networks are Deep Neural Networks (DNN) with convolutional and max-pooling layers.  
Convolutional Neural Networks (CNN) can be traced back to the Neocognitron \cite{fukushima:1980} in 1980. They were first successfully applied to relatively small tasks such as digit recognition \cite{Simard:2003}, image interpretation \cite{behnke:2003} and object recognition \cite{lecun:1998}. Back then their size was greatly limited by the low computational power of available hardware. Since 2010, however, DNN have greatly profited from Graphics Processing Units (GPU). Simple GPU-based multilayer perceptrons (MLP) establised new state of the art results \cite{Ciresan:2010} on the MNIST handwritten digit dataset~\cite{lecun:1998} when made both deep and large
(augmenting the training set by artificial samples helped to avoid overfitting).
2011 saw the first 
implementation \cite{Ciresan:2011a} of GPU-based DNN on the CUDA parallel computing platform~\cite{NVIDIA:2009}. This yielded new
benchmark records on multiple object detection tasks.
The field of Deep Learning with Neural Networks exploded.
Multi-Column DNN \cite{Ciresan:2012b} improved previous results by over 30\% on many benchmarks including: handwritten digits (MNIST)~\cite{lecun:1998} and Latin letters (NIST SD 19)~\cite{NIST}; Chinese characters (CASIA)~\cite{cheng-lin:2010}; traffic signs (GTSRB)~\cite{stallkamp:2011}; natural images (CIFAR 10)~\cite{krizhevsky:2009}.  Another flavor of DNN~\cite{Krizhevsky:2012} greatly improved the accuracy on a subset of ImageNet \cite{ILSVRC:2012}.  Recently, Google parallelized a large DNN on a cluster with over 10000 CPU cores~\cite{Dean:2012}.

For image \emph{classification}, the DNN returns a vector of class posterior probabilities when provided with an input patch whose fixed width and height usually does not exceed a few hundreds of pixels and depends on the network architecture. But DNN also excel at image segmentation and object detection \cite{Ciresan:2012f}.
For \emph{segmentation}, image data within a square patch of odd size is used to determine the class of its central pixel.  The network is trained on patches extracted from a set of images with ground truth segmentations (i.e. the class of each pixel is known).  To segment an unseen image, the trained net is used to classify all of its pixels. Object
\emph{detection} within an image is trivially cast as a segmentation problem: pixels close to the centroid of each object are classified differently from background pixels.  Once an unseen image is segmented, the centroid of each detected object is determined using simple image processing techniques.

Solving segmentation and detection tasks requires to apply the network to every patch contained in the image, which is prohibitively expensive when implemented in the naive, straightforward way.  Consider a net with a convolutional layer immediately above the input layer: when evaluating the first patch contained in the input image, the patch is convolved with a large number of kernels to compute the output maps; when evaluating the next (typically overlapping) patch, such convolutions are re-evaluated -- a huge amount of redundant computation.  It is better to compute each convolution only once \emph{for the whole input image}: the resulting set of images (which we will refer to as \emph{extended maps}) contain the maps for each patch contained in the input image.

In the particular case of a CNN {\bf without} max-pooling layers, this optimization is trivially implemented by computing all convolutions in the first layer on the entire input image, then computing all convolutions in subsequent layers on the resulting extended maps.  This approach \cite{farabet-suml-11}  
yielded real time detection performance~\cite{farabet-ecv-09} when combined with dedicated FPGA or even ASIC integrated circuits.

However, present DNN owe much of their power to max-pooling layers interleaved with convolutional layers.  Max-pooling cannot be handled using the straightforward approach outlined above.  For example, when we perform a $2 \times 2$ max-pooling operation on an extended map, we obtain a smaller extended map which does \emph{not} contain information from all the patches contained in the input image; instead, only patches whose upper left corner lies at even coordinates of the original image are represented.  Any subsequent max-pooling layer would further aggravate the problem.

Our contribution consists in an optimized forward-propagation approach which avoids such problems by \emph{fragmenting} the extended maps resulting from each max-pooling layer, such that each fragment contains information independent of other fragments, and the union of all fragments contains information about all the patches in the input image.  A similar approach was previously used~\cite{DBLP:conf/caip/NasseTF09} for handling a single subsampling layer in a simple CNN for face detection.  Our mechanism, however, is completely general. It handles arbitrary architectures mixing convolutional and max-pooling layers in any order, and ensures that no redundant computation is performed at any stage.

\section{Method}

We consider nets composed by four types of layers \cite{Ciresan:2011a}: input, convolutional, max-pooling and fully-connected.  In the following, different layers are denoted by index $l$.  Nets are formed by an input layer ($l=0$), followed by a set of convolutional and max-pooling layers $l\in\{1, \hdots, L\}$, followed by a number of fully-connected layers.
The optimization described in this paper concerns convolutional and max-pooling layers, and allows to find the outputs of layer $L$.  We do not discuss forward-propagation in fully-connected layers, where a trivial approach does not suffer from redundant computations.

To simplify notation, we consider nets with square maps and square kernels, but this is trivially generalized to rectangular maps and filters.  Sets are denoted by bold symbols.

\begin{figure*}[bth!]
\begin{center}
\includegraphics[width=\textwidth]{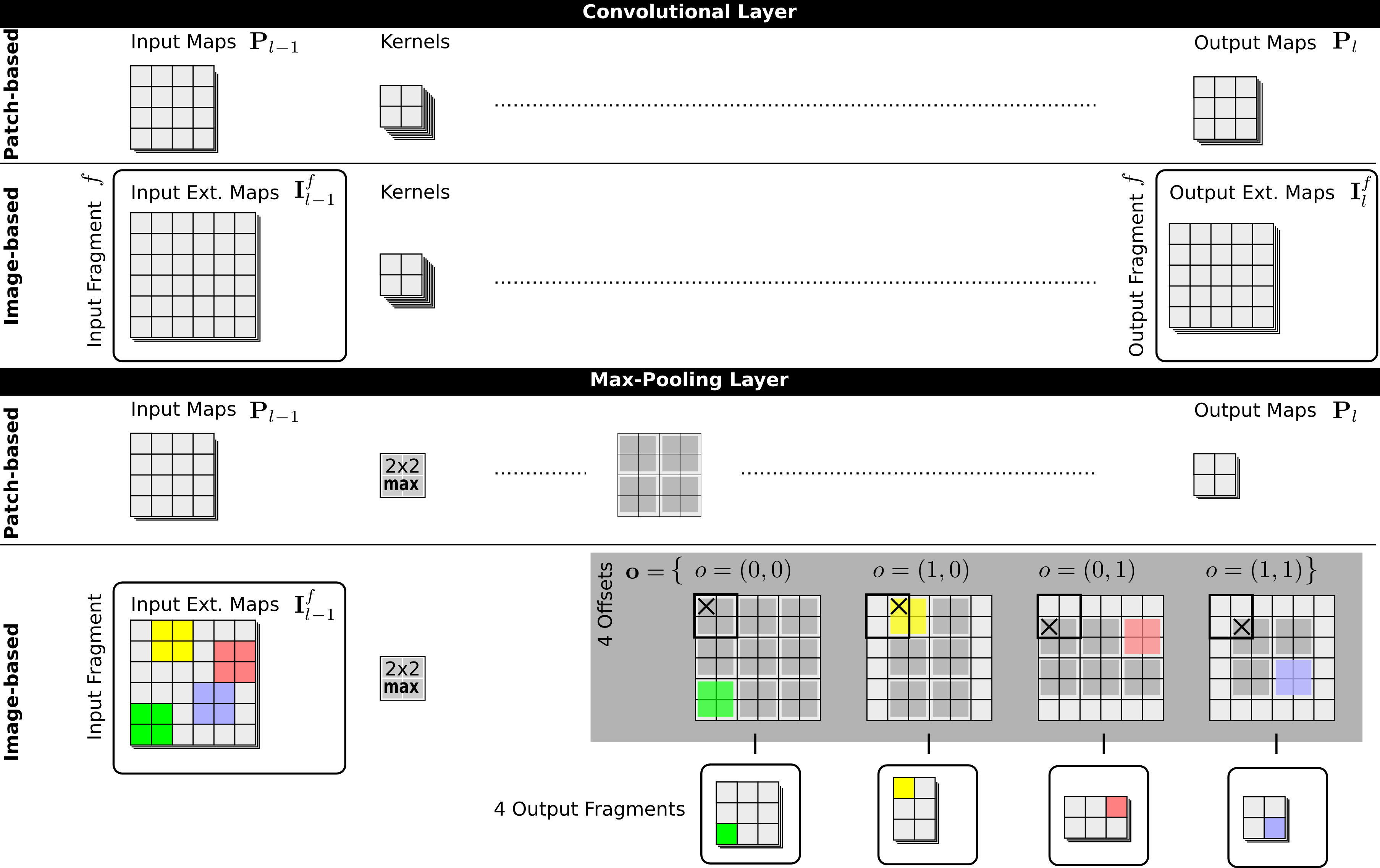}
\end{center}
\caption{Patch-based and image-based forward propagation techniques for convolutional and max-pooling layers.
\label{fig:illustration}} 
\end{figure*}

We first recall how convolutional and max-pooling layers are forward-propagated at the patch level (Section~\ref{ss:method:patch}). Then we extend the approach by the proposed optimization, which operates at the level of the whole image (Section~\ref{ss:method:image}).  Figure~\ref{fig:illustration} illustrates both approaches.

\subsection{Patch-level testing}\label{ss:method:patch}

The square input patch is represented as a set $\mathbf{P}_0$ containing one or more input maps (depending on the number of input channels). Let $w_0$ denote the width and height of such maps (i.e., the size of the input patch).
Because the input image and all kernels are assumed square, maps obtained as the output of any intermediate layer $l$ (i.e., contained in $\mathbf{P}_l$) will be square.

\subsubsection{Convolutional layers}
Let $l$ denote the index of a convolutional layer.  The layer's output is a set $\mathbf{P}_l$ of square maps with size $w_l$.
$\mathbf{P}_l$ is obtained as a function of $\mathbf{P}_{l-1}$ \cite{Ciresan:2011a}.  $w_l = w_{l-1}-(k-1)$, where $k$ is the width of the (square) kernels of layer $l$.  In general, the number of maps may change after a convolutional layer, i.e.: $\left|\mathbf{P}_l\right| \neq \left|\mathbf{P}_{l-1}\right|$.

\subsubsection{Max-pooling layers}
Let $l$ denote the index of a max-pooling layer.  The layer's output is a set $\mathbf{P}_l$ of square maps with size $w_l$.
$\mathbf{P}_l$ is obtained as a function of $\mathbf{P}_{l-1}$ \cite{Ciresan:2011a}.  $w_l = w_{l-1}/k$, where $k$ is the size of the square max-pooling kernel; the architecture of the net is such that $\text{mod}(w_{l-1},k)=0$.  The number of maps is unchanged after a max-pooling layer, i.e. $\left|\mathbf{P}_l\right| = \left|\mathbf{P}_{l-1}\right|$.

\subsection{Optimized testing on images}\label{ss:method:image}

Let us now consider a square input image of size $s \geq w_0$.  We want to compute the network outputs on all windows completely contained within it -- i.e. $(s-w_0+1)^2$ windows (patches).

We represent the output data of a given layer $l$ of the net as a set containing $F_l$ \emph{fragments}.  We denote each fragment in layer $l$ by an index $f\in\{1,\hdots, F_l\}$.  Each fragment
$f\in\{1,\hdots, F_l\}$ 
is associated to a set $\mathbf{I}_l^f$ of \emph{extended maps}.

Extended maps in the same fragment all have the same size; extended maps in different fragments may have different sizes, and not all such sizes may be square even though the input image is.  Let ${s_{x,l}^f}$ and ${s_{y,l}^f}$ denote width and height of the extended maps in set $\mathbf{I}_l^f$,  respectively.


The input image is provided as a single fragment, therefore $F_0 = 1$.  Such fragment contains a set $\mathbf{I}_0^1$ of square extended maps with sizes
\begin{eqnarray}
s_{x,0}^1&=&s\\
s_{y,0}^1&=&s\;.
\end{eqnarray}

\subsubsection{Convolutional layers}

Let $l$ denote the index of a convolutional layer.  Its output consists of a set of fragments, such that $F_l = F_{l-1}$.

A given fragment $f\in\{1,\hdots, F_l\}$ contains a set $\mathbf{I}_l^f$ of extended maps; each  has size
\begin{eqnarray}
s_{x,l}^f &=& s_{x,l-1}^f-(k-1)\\
s_{y,l}^f &=& s_{y,l-1}^f-(k-1)\;,
\end{eqnarray}
where $k$ is the size of the (square) kernels of layer $l$.  Again, note that maps in different fragments may have different sizes, but all maps in the same fragment have the same size.

$\mathbf{I}_l^f$ is obtained as a function of $\mathbf{I}_{l-1}^f$, using the same operations as with patch-level forward propagation. However, convolutions are performed on the (large) extended maps rather than on small maps, like in the patch-level approach.

\subsubsection{Max-pooling layers}

Let $l$ denote the index of a max-pooling layer.  Its output consists of a set of fragments, such that $F_l = k^2 F_{l-1}$ fragments, where $k$ is the size of the square max-pooling kernel.  In particular, each input fragment $f_\text{in} \in \{1, \hdots, F_{l-1}\}$ generates $k^2$ output fragments.

Consider a given input fragment $f_\text{in}$, associated with the set $\mathbf{I}_{l-1}^{f_\text{in}}$ containing the input extended maps.  Let $\mathbf{o}$ be a set of $k^2$ 2D offsets defined as the Cartesian product $\{0, 1, \hdots, k-1\} \times \{0, 1, \hdots, k-1\}$.  E.g., for $k=2$: $$\mathbf{o}=\{(0,0),(1,0),(0,1),(1,1)\}\;.$$

For a given input fragment $f_\text{in}$ and for each offset $o \in \mathbf{o}$, $o=(o_x, o_y)$, one output fragment $f$ is produced.  Each of the extended maps in $f$ is generated by applying the max-pooling operation to the corresponding extended map in $f_\text{in}$, by starting at the top left offset $(x,y) = o$.
Specifically, the pixel at coordinates $(\bar{x},\bar{y})$ in the output map is computed as the maximum of all pixels in the corresponding input map at coordinates $(x,y)$ such that:
\begin{equation}
\begin{array}{rcccl}
o_x+k\bar{x} &\leq& x &\leq& o_x+k\bar{x}+k-1\\
o_y+k\bar{y} &\leq& y &\leq& o_y+k\bar{y}+k-1\;.
\end{array}
\end{equation}

Then the size of the extended maps in $\mathbf{I}_l^f$ is given by:
\begin{eqnarray}
s_{x,l}^f &=& \text{div}\left(\left(s_{x,l-1}^{f_\text{in}}-o_x\right),k\right)\\
s_{y,l}^f &=& \text{div}\left(\left(s_{y,l-1}^{f_\text{in}}-o_y\right),k\right)\;,
\end{eqnarray}
where $\text{div}$ denotes the integer division operation.  The max-pooling operation thus ignores the following parts of the input extended maps:
\begin{itemize}
 \item $o_y$ leftmost columns;
 \item $o_x$ top rows;
 \item $\text{mod}\left(\left(s_{x,l-1}^{f_\text{in}}-o_x\right),k\right)$ rightmost columns;
 \item $\text{mod}\left(\left(s_{x,l-1}^{f_\text{in}}-o_x\right),k\right)$ bottom rows.
\end{itemize}

\section{Discussion and Results}

Convolutional layers do not alter the number of fragments (and operate on each fragment independently), whereas each max-pooling layer produces $k^2$ times the number of fragments given at its input.  Therefore, the final number of fragments generated by a net is equal to the product of the squares of the kernel sizes of all its max-pooling layers; for example, the net in Table~\ref{tab:net_architecture} produces $2^2 \cdot 2^2 \cdot 2^2 \cdot 2^2 = 256$ fragments at the output of layer 8.
Note that for all layer types, including fully-connected layers, data in a fragment at layer $l$ only depends on data in a single fragment at layer $l-1$.

Let $w_l$ be the size of the map at layer $l$ when using the patch-based approach.  Now consider our image-based approach, and the set $\mathbf{I}_l^f$ of extended maps at a given fragment $f$ for layer $l$.  Any $w_l \times w_l$ subimage cropped from such extended maps corresponds to the contents of the corresponding maps for some patch contained in the original image.  A single fragment contains data for a subset of the patches contained in the original image.  Collectively, all fragments at a given layer contain data for the whole set of all patches contained in the original image.

\begin{table}[bth!]
\caption{11-layer architecture for network N4 used in \cite{Ciresan:2012f}.
\label{tab:net_architecture}} 
\begin{center}
\small
\begin{tabular*}{\columnwidth}{@{\extracolsep{\fill}}c l l c}
\toprule
Layer		&	Type		&	Maps ($\left|\mathbf{P}_{l}\right|$)	&	Kernel 	\\
($l$)		&				&	  and neurons 					&	($k_l\times k_l$)		\\
\midrule
0		&	input			&	1 map of 95x95 neurons		&			\\
1		&	convolutional	&	48 maps of 92x92 neurons		&	4x4		\\
2		&	max pooling	&	48 maps of 46x46 neurons		&	2x2		\\
3		&	convolutional	&	48 maps of 42x42 neurons		&	5x5		\\
4		&	max pooling	&	48 maps of 21x21 neurons		&	2x2		\\
5		&	convolutional	&	48 maps of 18x18 neurons		&	4x4		\\
6		&	max pooling	&	48 maps of 9x9 neurons		&	2x2		\\
7		&	convolutional	&	48 maps of 6x6 neurons		&	4x4		\\
8		&	max pooling	&	48 maps of 3x3 neurons		&	2x2		\\
9		&	fully connected	&	200 neurons				&	1x1		\\
10		&	fully connected	&	2 neurons					&	1x1		\\
\bottomrule
\end{tabular*}
\end{center}
\end{table}

\subsection{Theoretical speedup}
\label{section:theoretical_speedup}

\begin{table*}[th!]
\caption{Theoretically required FLOPS for convolutional layers when segmenting a $512 \times 512$ image using patch-based ($\text{FLOPS}_l^\text{patch}$) and image-based ($\text{FLOPS}_l^\text{image}$) approaches.  Net architecture in Table~\ref{tab:net_architecture}. See text for details.
\label{tab:flops}} 
\begin{center}
\small
\begin{tabular*}{\textwidth}{@{\extracolsep{\fill}}lrrrrrrrrrr}
\toprule
Layer ($l$)		&	$s$ 	&	$s_{l-1}$	&	$\left|\mathbf{P}_{l-1}\right|$	&	$\left|\mathbf{P}_{l}\right|$	&	$w_l$	&	$k_l$	&	$F_l$	&	$\text{FLOPS}_l^\text{patch} [\cdot 10^9]$ 	&	$\text{FLOPS}_l^\text{image} [\cdot 10^9]$	&	speedup\\
\midrule
1			&	512	&	559	&	1									&	48								&	92		&	4	&	1	&	3408	&	0.5	&	7114.8\\
3			&	512	&	279	&	48									&	48								&	42		&	5	&	4	&	53271	&	35.9	&	1485.1\\
5			&	512	&	139	&	48									&	48								&	18		&	4	&	16	&	6262	&	22.8&	274.7\\
7			&	512	&	69	&	48									&	48								&	6		&	4	&	64	&	695	&	22.5&	30.9\\
\midrule
Total		&		&										&									&			&			&	&	&	63636	&	81.6	&	779.8\\	
\bottomrule
\end{tabular*}
\end{center}
\end{table*}

We now discuss the speedup of our image-based approach in comparison to separate evaluation of all patches contained in the input image.
We consider as an example the largest network (Table~\ref{tab:net_architecture}) used in \cite{Ciresan:2012f} for neuronal membrane segmentation~\cite{website:competition}. The image size (one slice with neuronal tissue data) is $512\times512$ pixels (see Figure~\ref{fig:challenge}). Its edges are mirrored, to get enough pixels for applying the network to all positions.  We limit our analysis to convolutional layers, which are by far the most computationally intensive part of a DNN.  Conversely, max-pooling layers are simple and fast, requiring less than 1\% of the computing time in most practical DNN.

For the patch-based approach, the required amount of floating-point operations (FLOPS) for computing the convolutions in layer $l$ when scanning an image by a DNN obeys the following formula: \\
$$\text{FLOPS}^\text{patch}_l = s^2 \cdot \left|\mathbf{P}_{l-1}\right| \cdot \left|\mathbf{P}_{l}\right| \cdot  w_l^2  \cdot k_l^2 \cdot 2,$$
 where $s^2$ is the number of pixels in the input image, and, for each convolutional layer $l$, $\left|\mathbf{P}_{l}\right|$ denotes the number of maps, $w_l^2$ the number of pixels of the map, and $k_l^2$ the number of kernel pixels.  The factor "2" reflects that we have one addition and one multiplication for each component of the dot product.

For the image-based approach, the FLOPS can be computed using the following formula: 
$$\text{FLOPS}^\text{image}_l={s_{x,l}} \cdot {s_{y,l}} \cdot \left|\mathbf{P}_{l-1}\right| \cdot \left|\mathbf{P}_{l}\right| \cdot F_l \cdot k^2_l \cdot 2,$$ where $s_{x,l} \cdot s_{y,l}$ represents the size of a fragment in layer $l$ (to simplify the computation, we assume all fragments have the same size, although they may differ in size by at most one pixel).
For the input layer ($l=0$), mirroring the borders implies that $s_{x,0}=s_{y,0}=s+(w_0-1)/2$.

Table~\ref{tab:flops} reports such computations for all convolutional layers in the network of Table~\ref{tab:net_architecture}.  The patch-based approach requires 779.8 times more FLOPS than the image-based approach.

\subsection{Experimental speedup}

In Table~\ref{tab:speed} we report computation times of the DNN in Table~\ref{tab:net_architecture}, when used to segment a $512 \times 512$ image using three different implementations:
\begin{description}
 \item[matlab-patch] a plain MATLAB implementation of the patch-based approach;
 \item[GPU-patch] a heavily optimized implementation of the patch-based approach running on a GTX-580 graphics card using CUDA; 
 \item[matlab-image] a plain MATLAB implementation of the image-based approach.
\end{description}

\begin{table}[bth!]
\caption{Speed for segmenting a $512 \times 512$ image using the net in table~\ref{tab:net_architecture}.
\label{tab:speed}} 
\begin{center}
\small
\begin{tabular*}{\columnwidth}{@{\extracolsep{\fill}}lrr}
\toprule
Method		&	Time per		&	Speedup relative\\
		&	image [s]		&   	to GPU-patch\\
\midrule
matlab-patch	&	24641.54		&	-			\\
GPU-patch	&	492.83			&	1			\\
matlab-image	&	15.05			&	32.8			\\
\bottomrule
\end{tabular*}
\end{center}
\end{table}

Results clearly show that the image-based implementation yields a dramatic speedup over patch-based approaches.  In particular, \emph{matlab-image} yields a $32$-fold speedup when compared to the highly-optimized \emph{GPU-patch} implementation, despite the former being implemented in a slower environment and without attention to low-level optimizations.  The impact of GPU and low-level optimizations is obvious as the \emph{GPU-patch} approach is $50$ times faster than \emph{matlab-patch}.

\section{Conclusions}

\begin{figure}[bth!]
\centering
\includegraphics[width=\columnwidth]{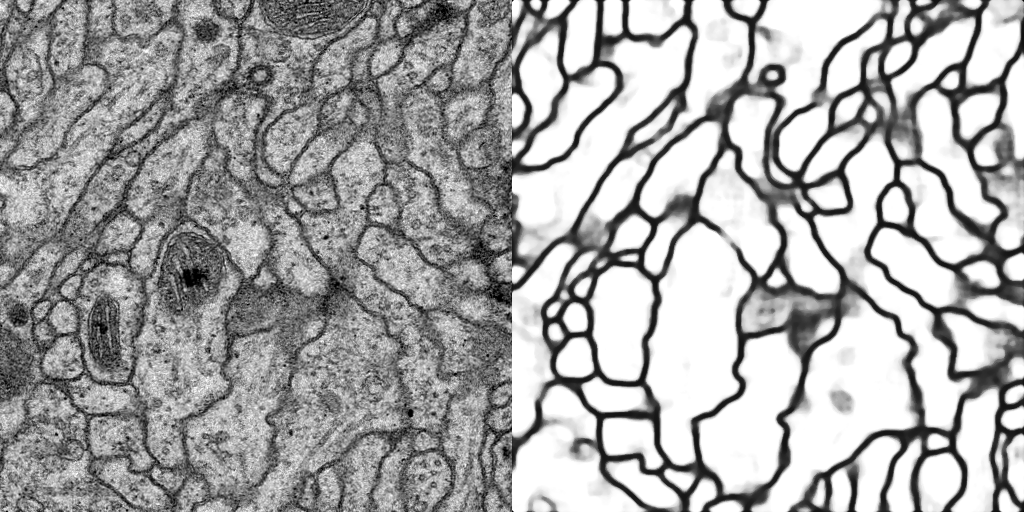}
\caption{An input electron microscopy slice (left) and the corresponding segmentation output (right).  Data from the ISBI EM segmentation challenge~\cite{website:competition}.  As our approach is an exact method, patch-based and image-based approaches yield identical results.
\label{fig:challenge}} 
\end{figure}

We greatly sped up forward-propagating deep neural networks on sliding windows. Our approach handles the complications due to max-pooling layers interleaved with convolutional layers, avoiding unnecessary computations. This is important for fast object detection and image segmentation. For huge nets such as those winning the ISBI Electron Microscopy Segmentation Challenge~\cite{website:competition,10.1371/journal.pbio.1000502} (see Figure~\ref{fig:challenge}), our approach is in theory almost three orders of magnitude faster than a straightforward patch-based forward-propagation approach. In practice, a simple MATLAB implementation yields a 32-fold speedup over a highly optimized patch-based GPU implementation.

\section{Acknowledgment}

This work was partially supported by the \emph{Supervised Deep / Recurrent Nets} SNF grant, Project Code 140399.

\bibliographystyle{IEEEbib}
\bibliography{bib}

\end{document}